\documentclass[10pt,twocolumn,letterpaper]{article}

\usepackage{iccv}
\usepackage{times}
\usepackage{epsfig}
\usepackage{graphicx}
\usepackage{amsmath}
\usepackage{amssymb}
\usepackage{array}
\usepackage{multirow}
\usepackage{tablefootnote}
\usepackage{pifont}
\usepackage[breaklinks=true,bookmarks=false]{hyperref}

\iccvfinalcopy % *** Uncomment this line for the final submission

\def\iccvPaperID{****} % *** Enter the ICCV Paper ID here

\ificcvfinal\fi

\begin{document}

%%%%%%%%% TITLE
\title{PERT: A Progressively Region-based Network for Scene Text Removal}

\author{Yuxin Wang, Hongtao Xie, Shancheng Fang, Yadong Qu and Yongdong Zhang\\
University of Science and Technology of China\\
{\tt\ \{wangyx58,qqqyd\}@mail.ustc.edu.cn, \{htxie,fangsc,zhyd73\}@mail.ustc.cn}
}

\maketitle
\ificcvfinal\thispagestyle{empty}\fi

%%%%%%%%% ABSTRACT
\begin{abstract}

Scene text removal (STR) contains two processes: text localization and background reconstruction. Through integrating both processes into a single network, previous methods provide an implicit erasure guidance by modifying all pixels in the entire image. However, there exists two problems: 1) the implicit erasure guidance causes the excessive erasure to non-text areas; 2) the one-stage erasure lacks the exhaustive removal of text region. In this paper, we propose a ProgrEssively Region-based scene Text eraser (PERT), introducing an explicit erasure guidance and performing balanced multi-stage erasure for accurate and exhaustive text removal. Firstly, we introduce a new region-based modification strategy (RegionMS) to explicitly guide the erasure process. Different from previous implicitly guided methods, RegionMS performs targeted and regional erasure on only text region, and adaptively perceives stroke-level information to improve the integrity of non-text areas with only bounding box level annotations. Secondly, PERT performs balanced multi-stage erasure with several progressive erasing stages. Each erasing stage takes an equal step toward the text-erased image to ensure the exhaustive erasure of text regions. Compared with previous methods, PERT outperforms them by a large margin without the need of adversarial loss, obtaining SOTA results with high speed (71 FPS) and at least 25\% lower parameter complexity. Code is available at \url{https://github.com/wangyuxin87/PERT}.

\end{abstract}
\section{Introduction}

As one of the most important mediums in information interaction, scene text contains quite a lot of sensitive and private information \cite{karatzas2015icdar,shi2016end,cheng2017focusing,li2019show}. To prevent these private messages from being used in illegal ways, Scene Text Removal (STR) task aims to remove the texts in the scene images and fill in the background information correspondingly. Benefiting from the development of Generative Adversarial Networks (GANs) \cite{karras2017progressive,karras2019style}, recent STR methods achieve promising results with various solutions \cite{isola2017image,zhang2019ensnet}. However, there are still two problems to be solved.

%  The main challenge in STR task is how to ensure the visual integrity of the overall image.
% Compared with manually modifying pictures by Photoshop editors, computer-based automatic scene text erasing is more labor-saving, which has important research significance.

\begin{figure}[t!p]
  \centering
  \includegraphics[width=8.2cm]{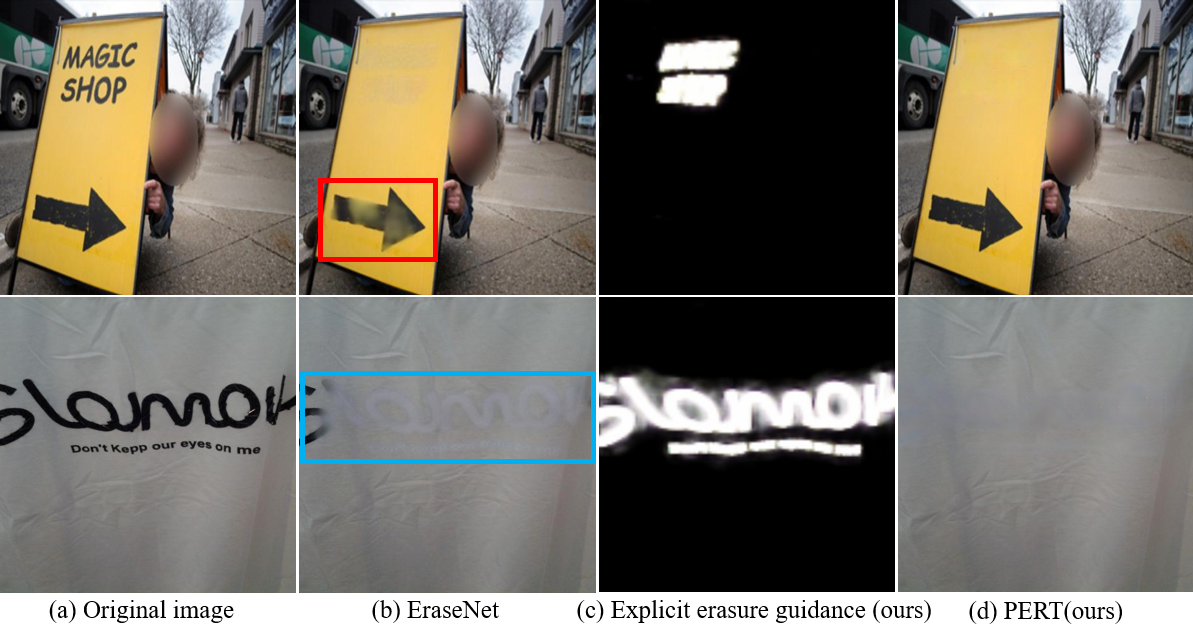}
  %\vspace{-1em}
  \caption{The comparison between PERT and EraseNet \cite{liu2020erasenet}. Compared with previous methods, PERT effectively handles the excessive erasure problem (red box) and inexhaustive erasure problem (blue box). Bounding box level annotations are used to supervise the explicit erasure guidance.}\label{fig:intro}
  % Only bounding box level annotations are used to supervise TLN.}\label{fig:intro}
  % \vspace{-1.5em}
\label{fig:significant}
\end{figure}

%  PERT provides an explicit erasure guidance to prevent the modification of non-text areas.

The first problem is the excessive erasure problem, which makes the poor integrity of non-text regions. Recent methods \cite{zhang2019ensnet,isola2017image} simply use paired images to train the model, integrating text localization and background reconstruction into a single network. However, since the text instances are sparse and exist in partial areas of scene images, such implicit erasure guidance that modifies all pixels in the entire image is not suitable for STR task. Though EraseNet \cite{liu2020erasenet} additionally uses a mask branch to enhance the text location perceiving, the pixel-wise reconstruction on the entire image is still performed under an implicit erasure guidance. As shown in red box of Fig. \ref{fig:intro} (b), such implicit erasure guidance has limited capability to maintain the integrity of non-text areas. In this paper, we are \textbf{the first} to argue that \emph{the explicit erasure guidance is the key}, which provides targeted and regional erasure on only text regions to prevent modifying non-text areas.

% Thus, these methods usually suffer from the excessive erasure problem of non-text areas.

%
% We argue that the incomplete decoupling of TL and BR has limited capability to provide accurate guidance during erasing. The complete decoupling contains two parts: structure decoupling and computing decoupling.

% is one of the core problem in STR task, which is \textbf{first} discussed in this paper and not be fully noticed by previous methods.

The second problem is the inexhaustive erasure problem, resulting in the remnants of text traces. Early methods \cite{nakamura2017scene,zhang2019ensnet} achieve text removal through a one-stage erasure, which has limited capability to obtain exhaustive erasure of text regions. Though recent methods \cite{tursun2020mtrnet++,liu2020erasenet} design a multi-stage eraser to refine the coarse erased image, due to the different learning difficulty caused by the same text-erased image supervision, it is difficult to balance the network architecture between the coarse and refinement stage. Thus, such imbalanced multi-stage erasure will leave some traces of text regions in the removal result (blue box of Fig. \ref{fig:intro} (b)). Based on the above analyses, how to construct the balanced multi-stage erasure needs to be explored.

% In addition, the stacked structure of multi-stage eraser significantly increases the parameter size.

% and achieve exhaustive erasure with a light network

% To improve the erasure performance of text regions, recent methods \cite{tursun2020mtrnet++,liu2020erasenet} design a multi-stage eraser to refining the coarse erased image. However, due to the different learning difficulty caused by the same text-erased image supervision, it is difficult to balance the network architecture between the coarse and refinement stage. It is worth noting that such imbalanced problem is not received enough attention by previous methods. As shown in blue box of Fig. \ref{fig:intro}, traces of the text region still remain in the erasure result. In addition, the stacked structure of multi-stage eraser contains a large parameter size and is difficult to be implemented. Thus, how to balance the learning difficulty and network structure among different erasing stages, and achieve exhaustive erasure with a light network is necessary for multi-stage eraser.

% Though recent multi-stage erasers \cite{tursun2020mtrnet++,liu2020erasenet} obtain promising results by refining the coarse erased image,

% Early methods \cite{nakamura2017scene,zhang2019ensnet} achieve text removal through a one-stage erasure, which are limited to generate exhaustive erasure of text regions. Thus,

\begin{figure}[t!p]
  \centering
  \includegraphics[width=8.8cm]{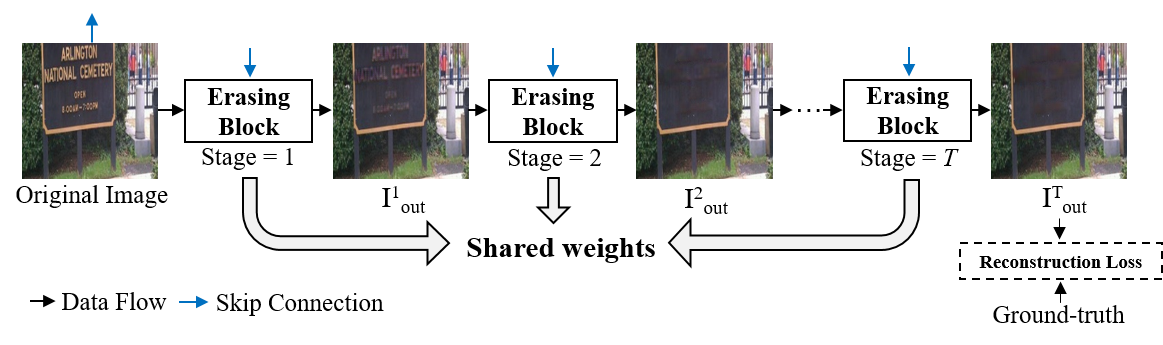}\\
  %\vspace{-1em}
  \caption{The pipeline of PERT. $I^t_{out}$ is the erased image from $t^{th}$ erasing stage.}
  \label{fig:pipeline}
  %\vspace{-1.5em}
\end{figure}

\begin{figure*}[t!p]
  \centering
  \includegraphics[width=16cm]{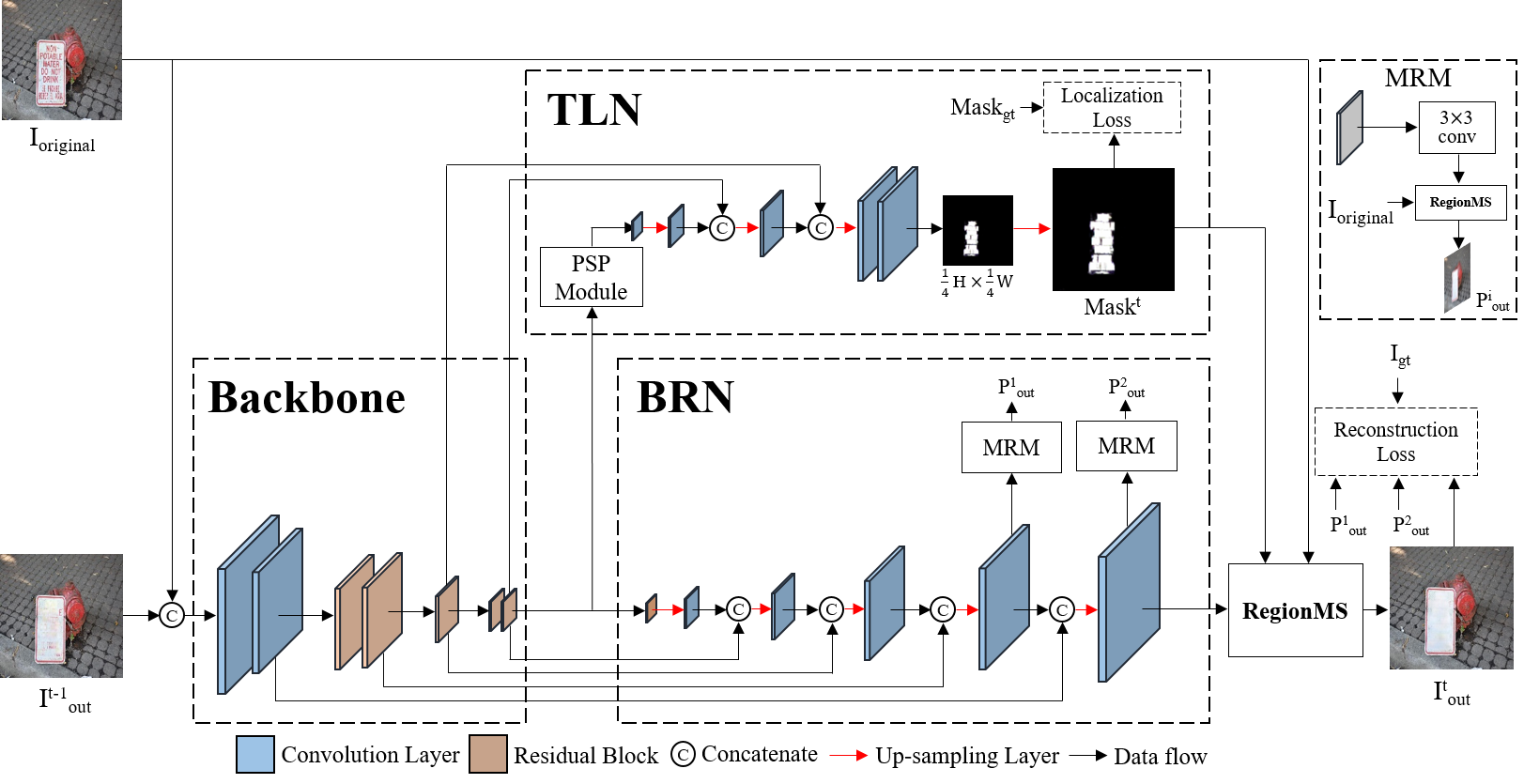}\\
  %\vspace{-1em}
  \caption{The architecture of erasing block. $I^{t}_{out}$ means the output from stage $t$, which has the same resolution of the original image $I_{original}$. $Mask^t$ is the mask map generated from TLN in stage $t$. MRM means Multi-scale Reconstruction Module.}\label{fig:block}
  %\vspace{-1.8em}
\label{fig:block_1}
\end{figure*}

In this paper, we propose a novel \emph{ProgrEssively Region-based scene Text eraser} (PERT) to handle above two problems from following two aspects: introducing an explicit erasure guidance and constructing the balanced multi-stage erasure. As shown in Fig. \ref{fig:pipeline}, PERT consists of several erasing blocks. Instead of integrating text localization and background reconstruction into a heavy network, we construct a lightweight decoupled structure of text localization network (TLN) and background reconstruction network (BRN) in each erasing block (shown in Fig. \ref{fig:block}). \textbf{1) The explicit erasure guidance}. As the text region predicted by TLN is a natural erasure guidance, we propose a new region-based modification strategy (RegionMS) to explicitly guide the BRN to only modify the predicted text regions. As background textures are directly inherited from the original image, RegionMS regards scene text removal as a targeted and regional erasure process to prevent the modification of non-text areas. Since the reconstruction learning on the final erased image aims to learn the stroke-level reconstruction rules, RegionMS enables TLN to adaptively perceive stroke-level information to further ensure the integrity of non-text areas (Fig. \ref{fig:intro} (c)) with only bounding box level annotations (Fig. \ref{fig:result} (d)). \textbf{2) The balanced multi-stage erasure}. To balance the network architecture and learning difficulty among different stages, we firstly construct a balanced erasure structure by sharing the parameters of each erasing block (shown in Fig. \ref{fig:pipeline}). Then, through only supervising the output of last erasing stage, PERT learns to adaptively balance the learning difficulty among different erasing stages, where each erasing stage aims to take an equal step toward text-erased image (shown in Fig. \ref{fig:iterative}). As parameters are shared among all erasing blocks, PERT is able to achieve exhaustive erasure with a light structure. In addition, to further improve the erasure performance, we propose a new Region-Global Similarity Loss (RG loss) to consider the feature consistency and visual quality of erasure results from both local and global perspective. Compared with previous methods, PERT obtains more exhaustive erasure of text regions while maintaining the integrity of non-text areas. Without the need of adversarial loss, PERT outperforms existing methods by a large margin with high speed (71 FPS) and at least 25\% lower parameter complexity.

The proposed method has several novelties and advantages: 1) To best of our knowledge, we are \textbf{the first} to propose an explicit erasure guidance in STR task. Furthermore, we also provide qualitative visualization to prove how it prevents the erasure of non-text regions and why it is more suitable for STR task. 2) Through designing a balanced erasure structure and supervising only the last erasing stage, PERT effectively balances the learning difficulty and network structure among different erasing stages, obtaining exhaustive erasure of text regions with a light architecture. 3) A new RG loss is proposed to improve the feature consistency and visual quality of erasure results. The SOTA results on both synthetic and real-world datasets demonstrate the effectiveness of our method.

%\vspace{-0.5em}
\section{Related Work}

% \subsection{Scene Text Removal}

% Scene text removal can be decomposed to two processes: text localization and background reconstruction.

Early STR methods \cite{khodadadi2012text,wagh2015text} mainly cascade the conventional text localization and background reconstruction processes for text removal. With the deep learning emerging as the most promising machine learning tool \cite{vaswani2017attention,zhan2019esir,Wan2019,qiao2020seed}, recent STR methods try to integrate the two independent processes into a single architecture by end-to-end training the network with paired images. Nakamura \emph{et al.} \cite{nakamura2017scene} implement a patch-based skip-connected auto-encoder for text removal. Under the premise of reducing the consistency of removal image, the coarse erasure result is obtained by taking the patch-level images as inputs. MTRNet \cite{tursun2019mtrnet} concatenates the text location map with original image to improve the erasure performance. As texts are sparse in the scene image \cite{tian2019learning,Wang_2019_CVPR,xue2019msr}, such implicit erasure guidance by modifying all the pixels in the entire image will cause the excessive erasure to non-text regions. Benefiting from the development of GANs \cite{brock2018large,zhu2017unpaired,zhang2017stackgan}, recent methods attempt to adopt adversarial loss to improve the erasure visuality. Though the adversarial loss significantly increases the training difficulty, the impressive improvement in visual quality makes it popular in STR task. Following the structure of cGAN \cite{mirza2014conditional}, Ensnet \cite{zhang2019ensnet} designs a local-aware discriminator to ensure the consistency of the erased regions. EraseNet \cite{liu2020erasenet} and MTRNet++ \cite{tursun2020mtrnet++} further construct a refinement-network to optimize the coarse output from the first erasure stage. Following these methods \cite{liu2020erasenet,tursun2020mtrnet++}, our method falls into multi-stage eraser, but we try to balance the learning difficulty and network structure among different stages. Compared with previous methods, PERT effectively handles the excessive and inexhaustive erasure problems from the aspects of introducing explicit erasure strategy and constructing balanced multi-stage erasure.

\section{Our Approach}
%\vspace{-0.5em}
\subsection{Pipeline}
The pipeline of PERT is shown in Fig. \ref{fig:pipeline}, which cascades $T$ lightweight erasing blocks. Through iteratively implementing erasing block on the erased image from previous stage, the output of the last erasing block is used as the final erasure result. To eliminate the effect of text characteristic losing, we concatenate the original image $I_{original}\in R^{H \times W \times 3}$ with $I^{t-1}_{out}\in R^{H \times W \times 3}$ ($H$ and $W$ are height and width respectively) to generate the input of $t^{th}$ erasing block (shown in Fig. \ref{fig:block}). For the first erasing stage, we concatenate $I_{original}$ with itself as the input.

% $I^{T}_{out}\in R^{H \times W \times 3}$ ($H$ and $W$ are height and width respectively)

%Finally, the erasure result from the last erasing stage is used as the final prediction.

 % $I^{t-1}_{out} \in R^{H \times W \times 3}$ at stage $t-1$

% In each erasing block (shown in Fig. \ref{fig:block_1}), we concatenate erased image $I^{t-1}_{out} \in R^{H \times W \times 3}$ from stage $t-1$ with the original image $I_{original} \in R^{H \times W \times 3}$ as input.

% In the implementation, we simply iteratively use the same erasing block in all stages. As the gradient will accumulate on the same erasing block, the simple \emph{loss.backward() \& optimizer.step()} are used for parameter updating.

% As shown in shown in Fig.\ref{fig:block_1}, at stage $t = 1$, we first concatenate the original image ($I_{original} \in R^{H \times W \times 3}$) with itself, and then we send it to the erase block. When stage $t > 1$, the input is the erased image $I^{t-1}_{out} \in R^{H \times W \times 3}$ from stage $t - 1$ concatenated with $I_{original}$. Finally, the erasure result from the last erasing stage is used as the final prediction. To adaptively balance the learning difficulty among different stages, we share the weights of each erasing block and only supervise the the final prediction from last stage.

%\vspace{-0.5em}
\subsection{Lightweight Erasing Block}

Inspired by the manual erasure process that firstly locating text regions and then performing regional modification, we design a decoupled structure in each erasing block. As shown in Fig. \ref{fig:block}, the erasing block contains a text localization network (TLN) for text location prediction and a background reconstruction network (BRN) for background textures reconstruction. Backbone network are shared by both TLN and BRN for feature extraction, which contains two convolution layers and five residual blocks \cite{he2016deep}.

%Then, we use a new region-based modification strategy (RegionMS) to perform targeted and regional erasure on the entire image (detailed in Sec. \ref{sec:regionms}).

% . In the backbone network, we firstly use a $4 \times 4$ convolution kernel with stride 2 to down-sample the input image. Then, five residual blocks \cite{he2016deep} are used to enhance the features.

% providing an explicit erasure guidance for local and targeted modification is theoretically more suitable for STR task, how to achieve such explicit erasure guidance is still a problem. Instead of integrating text localization and background reconstruction into a single network

% We set the down-sample ratio to 2 in the $2^{nd}$, $3^{rd}$ and $5^{th}$ residual block. In order to reduce the parameter size of the network, features extracted from backbone are shared between TLN and BRN.

% As the STR task consists of two phases (text localization and background reconstruction), instead of using a single network to integrate two independent phases, we design a parallel branches to decouple the erasure network to a lightweight TLN and a lightweight BRN. In TLN,

% TLN aims to predict text location information to explicitly guide the erasure process.

As text instances have large-variance scales, generating the robust representation of multi-scale texts is necessary for accurate scene text detection \cite{wang2019efficient,Wang2019}. Inspired by PSP Module \cite{zhao2017pyramid} which obtains promising performance in long-range dependency capture, we implement a PSP Module in TLN for robustly representing multi-scale texts. To reduce the computation cost, we only gradually up-sample the feature map to 1/4 size of the original image (bilinear interpolation is employed), and directly resize it to the same size of input image. Mask map $Mask^t \in [0,1]$ is generated through a sigmoid layer, where $t$ is the stage number. In the training stage, we supervise $Mask^t$ with the bounding box level annotations.

In order to learn the robust reconstruction rules of background textures, BRN is expected to contain two characteristics: 1) Modeling low-level texture information for background and foreground texture perception. 2) Capturing high-level semantics for feature representation enhancement. Thus, we construct BRN with skip connections to perceive low-level structures from shallow layers, and employ a relative deep network for high-level semantics capture. As the background reconstruction is a more challenging task than text localization, we use the deconvolution operation in the up-sampling layer to enhance the feature representation ability. Furthermore, the Multi-scale Reconstruction Module (MRM) is constructed to predict multi-scale erasure results ($P^1_{out}$ and $P^2_{out}$), which is only implemented in the training stage for performance boosting \cite{zhang2019ensnet}.

Instead of integrating the text localization and background reconstruction into an heavy network, our decoupled structure effectively reduces the parameter size, which only needs two lightweight network for localization and reconstruction respectively (detailed in Sec. \ref{sec:com}).

% Inspired by FPN \cite{lin2017feature}, which achieves impressive performance in these two aspects, we adopt a similar architecture to construct the BRN, while reducing the channels in up-sampling branches.

% By doing these, we successfully construct a decoupled structure of TLN and BRN.

% Through decomposing STR process to two sperate operations rather than integrating them in a single network, PERT successfully reduces the learning difficult (proved in Fig. \ref{fig:iterative}).

% However, there is not a specific erasure strategy to guide modification abandon the capability of text localization. Thus, it is necessary to provide a new erasure strategy for scene text removal

%\vspace{-0.5em}
\subsection{Region-based Modification Strategy}
\label{sec:regionms}

% Different from previous methods perform background reconstruction learning in the entire image, we provide an explicit erasure guidance for targeted and regional erasure, which prevent the modification of non-text areas and help BRN to learn target reconstruction rules of text regions.

%Though performing targeted and regional erasure by an explicit erasure guidance is theoretically

Different from previous implicit erasure guidance methods \cite{zhang2019ensnet,liu2020erasenet}, which modify all the pixels in the entire image, we provide an explicit erasure guidance for targeted and regional erasure. Such explicit erasure guidance is achieve by a region-based modification strategy (RegionMS).

% As the texts are sparse and exist in only regional areas of scene images, providing an explicit erasure guidance for targeted and regional erasure is theoretically more suitable for STR task, which avoids modifying the non-text areas. Based on the natural erasure guidance predicted by TLN, we introduce a new erasure strategy in this section, named region-based modification strategy (RegionMS).

%  However, how to achieve such explicit erasure process is still a problem.

RegionMS is formulated in Eq. \ref{eq:dfl}. Firstly, to sample the candidate erasure areas, we use the text mask map $Mask^{t}\in [0,1]$ to filter reconstructed image $I'^{t}$ from BRN. Thus, BRN will only modify the pixels in the predicted text regions. Relatively, to prevent the erasure of non-text areas, we preserve the background textures by performing element wise product between $(1 - Mask^{t})$ and original image $I_{original}$. It is worth mentioning that RegionMS is also employed in MRM to generate multi-scale predictions ($P^1_{out}$ and $P^2_{out}$), and the corresponding mask map is directly resized from $Mask^{t}$.

% We use the text localization map $Mask^{t}\in [0,1]$ to filter reconstructed image $I'^{t}$ from BRN and utilize the $(1 - Mask^{t})$ to maintain the non-text textures in the original image $I_{original}$. Without the complex architecture, RegionMS contains zero parameters. Specially, RegionMS is also used in MRM to generate multi-scale predictions ($P^1_{out}$ and $P^2_{out}$), where the corresponding mask map is directly resized from $Mask^{t}$.
%\vspace{-0.1em}
\begin{equation}\label{eq:dfl}
{
I^{t}_{out}= Mask^{t} \times I'^{t} + (1 - Mask^{t}) \times I_{original}
}
\end{equation}

To better illustrate the significance of RegionMS in ensuring the integrity of non-text areas, we provide some visualization of $Mask^{t}$. The details about this part is available in Sec. \ref{sec:stroke_qual}

\subsection{Balanced Multi-stage Erasure}

As the deep network can erase a large-range step for difficult cases while the shallow network provides a small-range one, due to the different learning difficulty caused by the same text-erased image supervision, it is difficult to balance the learning difficulty and network architecture in the multi-stage erasure \cite{tursun2020mtrnet++,liu2020erasenet}. Thus, we adopt the most straightforward approach by dividing the overall learning difficulty and total network structure into $T$ equal parts, where $T$ is an ablation study in Sec. \ref{sec:ab_study}.

Firstly, we construct a balanced erasure structure by implementing the erasing block with same structure in each erasing stage. To reduce the parameter complexity, we share the parameters among all erasing blocks. As the generation of text-erased image in $T$ different erasure degrees requires a lot of human costs, we simply supervise the output of only last erasing block, guiding the network to adaptively balance the learning difficulty among different stages. By doing these, a new balanced multi-stage erasure is obtained, where each erasing stage aims to take an equal step toward the text-erased image for exhaustive erasure (detailed in Sec.  \ref{sec:stroke_qual}).

\subsection{Training Objective}
\label{sec:training}

We use the original image $I_{original}$, text-removed ground-truth $I_{gt}$ and localization map $Mask_{gt}$ in bounding-box level (0 for non-text regions and 1 for text regions) to train PERT. The total loss contains two parts: localization loss and reconstruction loss. In order to adaptively balance the learning difficulty among different erasing stages, we only implement the reconstruction loss to the erased images ($I_{out}, P^1_{out}$ and $ P^2_{out}$) in the final erasing stage. In contrast, we supervise the mask map ($Mask^t$) in each erasing stage to guarantee an accurate erasure guidance.

\subsubsection{Localization loss} We use dice loss defined in \cite{wang2019dsrn} to guide the learning process of TLN. As shown in Eq. \ref{eq:mask}, $p_i$ is the prediction and $y_i$ is the ground-truth.
\begin{equation}\label{eq:mask}
{
L_{loc}= 1 - \frac{{2\sum\nolimits_i {{p_i}{y_i}} }}{{\sum\nolimits_i {{p_i} + } \sum\nolimits_i {{y_i}} }}
}
%\vspace{-0.1em}
\end{equation}

\subsubsection{Reconstruction loss} Benefiting from the RegionMS and balanced multi-stage erasure, the simple similarity losses are sufficient to train PERT.

1) Region-Global Similarity Loss (RG loss). The RG loss is newly proposed in this paper to consider the feature consistency and visual quality of erasure results from both local and global perspective. RG loss contains two parts (shown in Eq. \ref{eq:lg}): region-aware similarity (RS) loss and global-aware similarity (GS) loss.
%\vspace{-0.1em}
\begin{equation}\label{eq:lg}
{
L_{RG}= L_{RS} + L_{GS}
}
%\vspace{-0.5em}
\end{equation}
\begin{equation}
\begin{aligned}
L_{RS} = & \sum_{i=1}^2 \alpha_{i}\|(P^{i}_{out} - I_{i,gt}) * Mask_{i,gt}\|_{1} + \sum_{n=1}^2 \beta_{i}\|(P^{i}_{out}\\
& - I_{i,gt}) *(1 - Mask_{i,gt})\|_{1} + \alpha\|(I_{out} - I_{gt})\\
& * Mask_{gt}\|_{1} + \beta\|(I_{out} - I_{gt}) *(1 - Mask_{gt})\|_{1}
\end{aligned}
\label{eq:multi}
\end{equation}

As shown in Eq. \ref{eq:multi}, RS loss takes multi-scale predictions $P^{i}_{out}$ into consideration. $Mask_{i,gt}$ and $I_{i,gt}$ are generated by directly resizing the mask map $Mask_{gt}$ and text-erased image $I_{gt}$. The RS loss assigns the pixels in text region with a higher weight. To be specific, we set $\alpha,\alpha_1,\alpha_2=13,10,12$ and $\beta,\beta_1,\beta_2=2, 0.8, 1$ respectively.

Different from RS loss, GS loss aims to penalize the consistency and enhance the visual quality from a global view. We firstly down-sample $n=3$ activation maps $\phi_n(I_{out})\in R^{H_n \times W_n}$ from the $4^{th}, 9^{th}, 16^{th}$ layer of pre-trained VGG16 network to $F^{n}_{out}\in R^{S_m \times S_m}$ through max-pooling (the same process to $I_{gt}$). Inspired by the pair-wise Markov random field, which is widely used to improve the spatial labeling contiguity, we compute the pair-wise similarities between ground-truth and predicted features. Let $\gamma^{n}_{ij}$ denotes the similarity between the $j^{th}$ pixel and  $i^{th}$ pixel in feature $F^{n}$ (shown in Eq. \ref{eq:similarity}). $\gamma^{n,out}_{ij}$ means the similarity from $F^{n}_{out}$.
\begin{equation}\label{eq:similarity}
{
\gamma^{n}_{ij}= (F^{n}_{i})^TF^{n}_{j}/(\|F^{n}_{i}\|_{2}\|F^{n}_{j}\|_{2})
}
\end{equation}

% $\phi_n(I_{gt})\in R^{H_n \times W_n}$ and  $F^{n}_{gt}\in R^{S_m \times S_m}$

In our experiments, we set $S_m$ = 8, 4 and 1 when m = 1, 2 and 3, which means we calculate the pair-wise similarities in three different scales ($\gamma^{n,m}_{ij}, m=1,2,3$) for each feature $\phi_n(I)$. Finally, the GS loss is formulated in Eq. \ref{eq:global}. We choose the squared difference to compute the pair-wise similarity.
\begin{equation}\label{eq:global}
{
L_{GS}=\sum_{n=1}^3 \sum_{m=1}^3(\frac{1}{S_m \times S_m}(\gamma^{n,m,out}_{ij} - \gamma^{n,m,gt}_{ij})^2)
}
\end{equation}

2) Negative SSIM Loss \cite{wang2004image}. This loss is used to analyse the degradation of structural information:
\begin{equation}\label{eq:ssim}
{
L_{ssim}= -SSIM(I_{out},I_{gt})
}
\end{equation}

3) VGG Loss. Inspired by previous STR methods \cite{zhang2019ensnet,liu2020erasenet}, we also adopt VGG loss ($L_{vgg}$) to improve the erasure results. The details can be obtained in the previous methods \cite{zhang2019ensnet,liu2020erasenet}.

Finally, the total loss function is formulated in Eq. \ref{eq:total}.
\begin{equation}\label{eq:total}
{
L= L_{loc} + L_{RG} + L_{ssim} + L_{vgg}
}
\end{equation}

\section{Experiments}

\subsection{Datasets and Evaluation}

\subsubsection{Datasets} We conduct the experiments following the setup of \cite{liu2020erasenet}. We train PERT on the only official training images of SCUT-Syn \cite{zhang2019ensnet} or SCUT-EnsText \cite{liu2020erasenet}, and then evaluate the model on the corresponding testing sets, respectively. Details of these two datasets can be found in the previous works \cite{zhang2019ensnet,liu2020erasenet}.

\subsubsection{Evaluation} To comprehensively evaluate the erasure results of our method, we use both Image-Eval (PSNR, MSSIM, MSE, AGE, pEPs and pCEPs) and Detection-Eval (precision (P), recall (R), F-measure(F), TIoU-precision (TP), TIoU-recall (TR), TIoU-F-measure(TF)). Details about Image-Eval and Detection-Eval can be found in the previous works \cite{liu2019tightness,liu2020erasenet}. A higher PSNR and SSIM or lower MSE, AGE, pEPs, pCEPs, P, R, F, TP, TR and TF represent better results.

\subsection{Implementation Details}
Data augmentation includes random rotation with maximum degree of $10^{\circ}$ and random horizontal flip with a probability of 0.3 during training stage. PERT is end-to-end trained using Adam optimizer. The learning rate is set to 1e-3. The model is implemented in Pytorch and trained on 2 NVIDIA 2080Ti GPUs.

To share the parameters among different stages, we iteratively use the same erasing block in all stages. As the gradient will accumulate on the same erasing block, the simple \emph{loss.backward() \& optimizer.step()} are used for parameter updating. Details are available in our submitted code.

\subsection{Ablation Study}
\label{sec:ab_study}
\subsubsection{The region-based modification strategy}
As shown in Tab. \ref{tab:region}, through introducing an explicit erasure guidance, the proposed PERT significantly improves the erasure performance by 0.66, 0.33, 0.0002, 0.3931, 0.0016 and 0.0008 in PSNR, MSSIM, MSE, AGE, pEPs and pCEPs respectively. We attribute this remarkable improvement to two reasons: 1) the RegionMS provides targeted and regional modification on only text-region textures without changing pixels in non-text areas, ensuring the integrity of text-free regions. 2) The RegionMS reduces the learning difficulty of reconstruction process, helping BRN to focus on targeted reconstruction rules of text regions without considering non-text areas. The qualitative visualization are detailed in Sec. \ref{sec:stroke_qual}.

\subsubsection{The balanced multi-stage erasure}
As shown in Tab. \ref{tab:itera}, the one-stage erasure ($T=1$) has a limited capability to reconstruct background textures. When we increase the number of erasing stages step-by-step, the balanced multi-stage erasure increases the performance on all metrics, and the relative increases are 1.85, 0.59, 0.0003, 0.5361, 0.0092 and 0.0034 in PSNR, MSSIM, MSE, AGE, pEPs and pCEPs respectively. Benefiting from sharing parameters among all erasing blocks, PERT step-by-step refines the erasure result with ZERO parameter size increase.

% 2) The balanced erasure structure successfully achieves exhaustive erasure with the low parameter complexity.

% reduces the learning difficulty in each stage, and endows the model with the ability to refine the erasure results from the previous stage with ZERO parameter size increase. Benefiting from the progressive erasing, PERT achieves significant improvement on all metrics, and the relative increases are 1.85, 0.59, 0.0003, 0.5361, 0.0092 and 0.0034 in PSNR, MSSIM, MSE, AGE, pEPs and pCEPs respectively.

\subsubsection{The Region-Global similarity loss}
The RG loss penalizes the feature consistency and enhances the visual quality from both local and global views. As shown in Tab. \ref{tab:reloss}, the RG loss achieves the improvement by 0.52, 0.1, 0.0001, 0.1556, 0.0026 and 0.0020 in PSNR, MSSIM, MSE, AGE, pEPs and pCEPs respectively. Benefiting from the balanced multi-stage erasure and RegionMS, the simple similarity losses are sufficient to train PERT without the need of adversarial loss.

% As the models in 2nd row of Tab. \ref{tab:region}\&\ref{tab:reloss} have the same implementation, the performance is same.

%To evaluate the proposed method, we use the metrics leveraged in \cite{nakamura2017scene} to evaluate the quality of the final outputs.
%\vspace{-0.5em}
\subsection{Comparison with State-of-the-Art Methods}
\label{sec:com}
% \vspace{-0.5em}

The quantitative results on SCUT-EnsText dataset are shown in Tab. \ref{tab:real}. The state-of-the-art performance demonstrates that the proposed PERT outperforms existing methods on all metrics. Though EraseNet constructs a mask branch to enhance the perception of text appearance, the explicit erasure guidacne in PERT effectively improves the erasure performance and achieves a new state-of-the-art result in both Detection-Eval and Image-Eval. For further fair comparison, we carefully reimplement EraseNet \cite{liu2020erasenet} with the same training setting as ours, the improvement is also consistent. Compared with GAN-based methods \cite{isola2017image,zhang2019ensnet,liu2020erasenet}, our method obtains impressive performance with much simpler training objective, using only simple similarity losses to guide the learning process. In addition, we quantitatively compare the erasure results with recent approaches on SCUT-Syn dataset. As shown in Tab. \ref{tab:syn}, the proposed PERT obtains the dominant performance compared with both GAN-based \cite{isola2017image,zhang2019ensnet,liu2020erasenet} and GAN-free methods \cite{nakamura2017scene}.

\begin{table}[t!p]
   \begin{center}
   %\vspace{-1em}
	\resizebox{1.0\linewidth}{!}{
   \begin{tabular}{|c|c|c|c|c|}
      \hline
      \multirow{2}{*}{Model} & PSNR & MSE & \multirow{2}{*}{pEPs}& \multirow{2}{*}{pCEPs}\\
       & MSSIM & AGE & &   \\
      \hline
      \multirow{2}{*}{w/o RegionMS} & 32.59 &  0.0016&  \multirow{2}{*}{0.0152}& \multirow{2}{*}{0.0096}\\
       & 96.62&  2.5764& & \\
      \hline
      \multirow{2}{*}{w/ RegionMS} & \textbf{33.25} & \textbf{0.0014}& \multirow{2}{*}{\textbf{0.0136}}& \multirow{2}{*}{\textbf{0.0088}}\\
       & \textbf{96.95}&  \textbf{2.1833}& & \\
      \hline
   \end{tabular}}
   \end{center}
   %\vspace{-1.2em}
   \caption{Ablation studies about the RegionMS.}
   \label{tab:region}
   %\vspace{-0.5em}
   %\vspace{-0.8em}
\end{table}
\begin{table}[t!p]
\begin{center}
%\vspace{-0.5em}
\begin{tabular}{|c|c|c|c|c|c|c|}
\hline
 \multirow{2}{*}{T} & PSNR & MSE & \multirow{2}{*}{pEPs} & \multirow{2}{*}{pCEPs} &  Param\\
 & MSSIM &  AGE&  & & size \\
\hline
% &  &  &  &  &  & \\
\multirow{2}{*}{T = 1} &  31.40&  0.0017 & \multirow{2}{*}{0.0228} & \multirow{2}{*}{0.0122} & \multirow{2}{*}{14.0M}\\
 & 96.36&  2.7194&  & &  \\
\hline
\multirow{2}{*}{T = 2} &  32.72& 0.0016& \multirow{2}{*}{0.0160} &  \multirow{2}{*}{0.0104} & \multirow{2}{*}{14.0M}\\
&  96.83&  2.3750& & & \\
\hline
\multirow{2}{*}{T = 3} & \textbf{33.25} & \textbf{0.0014} & \multirow{2}{*}{\textbf{0.0136}} & \multirow{2}{*}{\textbf{0.0088}} &  \multirow{2}{*}{14.0M}\\
& \textbf{96.95}&  \textbf{2.1833}&  & & \\
% 4 &  \textbf{33.34}&  96.94&  0.0015&  2.2313&  0.0142& 0.0093\\
\hline
\end{tabular}
\end{center}
%\vspace{-1em}
%\vspace{-1.5em}
\caption{Ablation studies about the number of erasing stages. Param means parameter.}
\label{tab:itera}
%\vspace{-0.5em}
\end{table}
\begin{table}[t!p]
   \begin{center}
   %\vspace{-0.5em}
	\resizebox{1.0\linewidth}{!}{
   \begin{tabular}{|c|c|c|c|c|c|}
      \hline
      \multirow{2}{*}{Model} & PSNR & \multirow{2}{*}{MSE} &  \multirow{2}{*}{AGE} & \multirow{2}{*}{pEPs} & \multirow{2}{*}{pCEPs}  \\
       & MSSIM & &  &  &\\
      %\hline
      %\multirow{2}{*}{SSIM} & 32.00 &  \multirow{2}{*}{0.0018} & \multirow{2}{*}{2.6537} & \multirow{2}{*}{0.0196} &\multirow{2}{*}{0.0137}\\
      % & 96.71&  &  & &\\
      %\hline
      %\multirow{2}{*}{SSIM+RA} & 32.65 & \multirow{2}{*}{0.0016} & \multirow{2}{*}{2.3714} & \multirow{2}{*}{0.0162} &\multirow{2}{*}{0.0106}\\
      % & 96.76&  &  &  &\\
      \hline
       \multirow{2}{*}{w/o RG loss} & 32.73 & \multirow{2}{*}{0.0015} & \multirow{2}{*}{2.3389} & \multirow{2}{*}{0.0162} &\multirow{2}{*}{0.0108}\\
       & 96.85&  &  & &\\
      \hline
      \multirow{2}{*}{w/ RG loss} & \textbf{33.25} & \multirow{2}{*}{\textbf{0.0014}} & \multirow{2}{*}{\textbf{2.1833}} & \multirow{2}{*}{\textbf{0.0136}}& \multirow{2}{*}{\textbf{0.0088}}\\
      & \textbf{96.95}& &  & &\\
      \hline
   \end{tabular}}
   \end{center}
   %\vspace{-1em}
   \caption{Ablation studies about the RG loss.}
   \label{tab:reloss}
   %\vspace{-2em}
\end{table}
\begin{table*}[t!p]
\begin{center}
\begin{tabular}{|l|c|c|c|c|c|c|c|c|c|c|c|c|}
\hline
\multirow{2}{*}{Method} &
\multicolumn{6}{c|}{Image-Eval} &
\multicolumn{6}{c|}{Detection-Eval}\\
\cline{2-13}
& PSNR & MSSIM & MSE & AGE & pEPs & pCEPs & P& R& F& TP& TR& TF \\
\hline
Original images & - & - & - & - & - & - & 79.4& 69.5& 74.1& 61.4& 50.9& 55.7\\
Pix2Pix \cite{isola2017image}& 26.6993 & 88.56 & 0.0037 & 6.0860 & 0.0480 & 0.0227 & 69.7& 35.4& 47.0& 52.0& 24.3& 33.1\\
STE \cite{nakamura2017scene}& 25.4651 & 90.14 & 0.0047 & 6.0069 & 0.0533 & 0.0296 & \textbf{40.9}& 5.9& 10.2& \textbf{28.9}& 3.6& 6.4\\
EnsNet \cite{zhang2019ensnet}& 29.5382 & 92.74 & 0.0024 & 4.1600 & 0.0307 & 0.0136 & 68.7& 32.8& 44.4& 50.7& 22.1& 30.8\\
EraseNet \cite{liu2020erasenet}& 32.2976 & 95.42 & 0.0015 & 3.0174 & 0.0160 & 0.0090 & 53.2& 4.6& 8.5& 37.6& 2.9& 5.4\\
EraseNet* \cite{liu2020erasenet} &  32.0486&  95.47& 0.0015& 3.2751& 0.0169& 0.0098 & 58.3& 3.8& 7.1& 42.1& 2.3& 4.4\\
\textbf{PERT} & \textbf{33.2493} & \textbf{96.95} & \textbf{0.0014} & \textbf{2.1833} & \textbf{0.0136} & \textbf{0.0088} & 52.7& \textbf{2.9}& \textbf{5.4}& 38.7& \textbf{1.8}& \textbf{3.5}\\
\hline
\end{tabular}
\end{center}
%\vspace{-1em}
\caption{Comparisons between previous methods and proposed PERT on SCUT-EnsText. * means our reimplementation.}
\label{tab:real}
%\vspace{-0.5em}
\end{table*}
\begin{table*}[t!p]
\begin{center}
\begin{tabular}{|l|c|c|c|c|c|c|}
\hline
Method & PSNR & MSSIM & MSE & AGE & pEPs & pCEPs \\
\hline
% &  &  &  &  &  & \\
Pix2Pix \cite{isola2017image}& 26.76 & 91.08 & 0.0027 & 5.4678 & 0.0473 & 0.0244 \\
STE \cite{nakamura2017scene}& 25.40 & 90.12 & 0.0065 & 9.4853 & 0.0553 & 0.0347 \\
EnsNet \cite{zhang2019ensnet}& 37.36 & 96.44 & 0.0021 & 1.73 & 0.0069 & 0.0020 \\
EraseNet \cite{liu2020erasenet}& 38.32 & 97.67 & \textbf{0.0002} & 1.5982 & 0.0048 & \textbf{0.0004} \\
EraseNet* \cite{liu2020erasenet}& 37.70 & 97.34 & 0.0003 & 1.8044 & 0.0059 & 0.0009 \\
\textbf{PERT} & \textbf{39.40} & \textbf{97.87} & \textbf{0.0002} & \textbf{1.4149} & \textbf{0.0045} & 0.0006\\
\hline
\end{tabular}
\end{center}
%\vspace{-1em}
\caption{Comparisons between previous methods and proposed PERT on SCUT-Syn. * means our reimplementation.}
\label{tab:syn}
%\vspace{-1em}
\end{table*}

To qualitatively compare the erasure results, we visualize some examples in Fig. \ref{fig:result}. Specially, we choose the latest and the most related approach \cite{liu2020erasenet} for a detailed comparison. 1) \textbf{The integrity of non-text regions}. Though EraseNet \cite{liu2020erasenet} introduces a mask subnetwork to enhance the perception of text appearance, the implicit erasure guidance may modify the pixels belonging to non-text regions (the first row in Fig. \ref{fig:result}). Benefiting from the targeted and regional erasure provided by RegionMS, PERT modifies only text regions while ensuring the integrity of non-text areas. 2) \textbf{The exhaustive erasure of text regions}. Through balancing the learning difficulty and network structure among different erasing stages, PERT effectively achieves more exhaustive erasure by progressively taking an equal step toward text-erased images. As shown in the second row and (g) in Fig. \ref{fig:result}, PERT achieves exhaustive erasure and provides text-erased image with high visual quality.

% EraseNet has limited capability to exhaustively erase the traces of text regions. However,

% As it is difficult to balance the relationship between network architecture and learning difficulty among different stages, the capability of EraseNet \cite{liu2020erasenet} for exhaustive erasure is limited, which leaves some traces of the original text traces (the second and (g) in Fig. \ref{fig:result}). Benefiting from the balanced progressive erasure structure, PERT adaptively balances the learning difficulty among different stages, giving more exhaustive erasure results.

% and the reduction of learning difficulty by progressive structure

\subsubsection{Model size and speed} We compare the parameter size between PERT and existing methods in Tab. \ref{tab:modelsize}. We attribute our low parameter complexity to following two reasons: 1) the decoupled erasure structure reduces the network learning difficulty, which only needs two lightweight networks for detection and reconstruction without constructing a heavy network to achieve both functions simultaneously. 2) PERT shares the parameters among all erasing blocks. As shown in Tab. \ref{tab:modelsize}, PERT effectively reduces the model size by at least 25\% from existing multi-stage erasers \cite{tursun2020mtrnet++,liu2020erasenet} without performance decrease. The speed comparison is shown in Tab. \ref{tab:speed_table}. Without the need of adversarial loss, PERT effectively reduces the training time. In the testing stage, PERT also obtains the comparable speed and achieves real-time inference. Based on the above analyses, PERT obtains a better balance among the erasure performance, parameter complexity and inference speed.

% Though MTRNet++ \cite{tursun2020mtrnet++} and EraseNet \cite{liu2020erasenet} construct a multi-stage erasure to improve the erasure results, such stacked structure results in a high parameter complexity and is suboptimal for the real application.

\subsection{Quantitative Analysis}
\subsubsection{The significance of explicit erasure guidance}
\label{sec:stroke_qual}

\begin{figure*}[t!p]
  \centering
  \includegraphics[width=15.5cm]{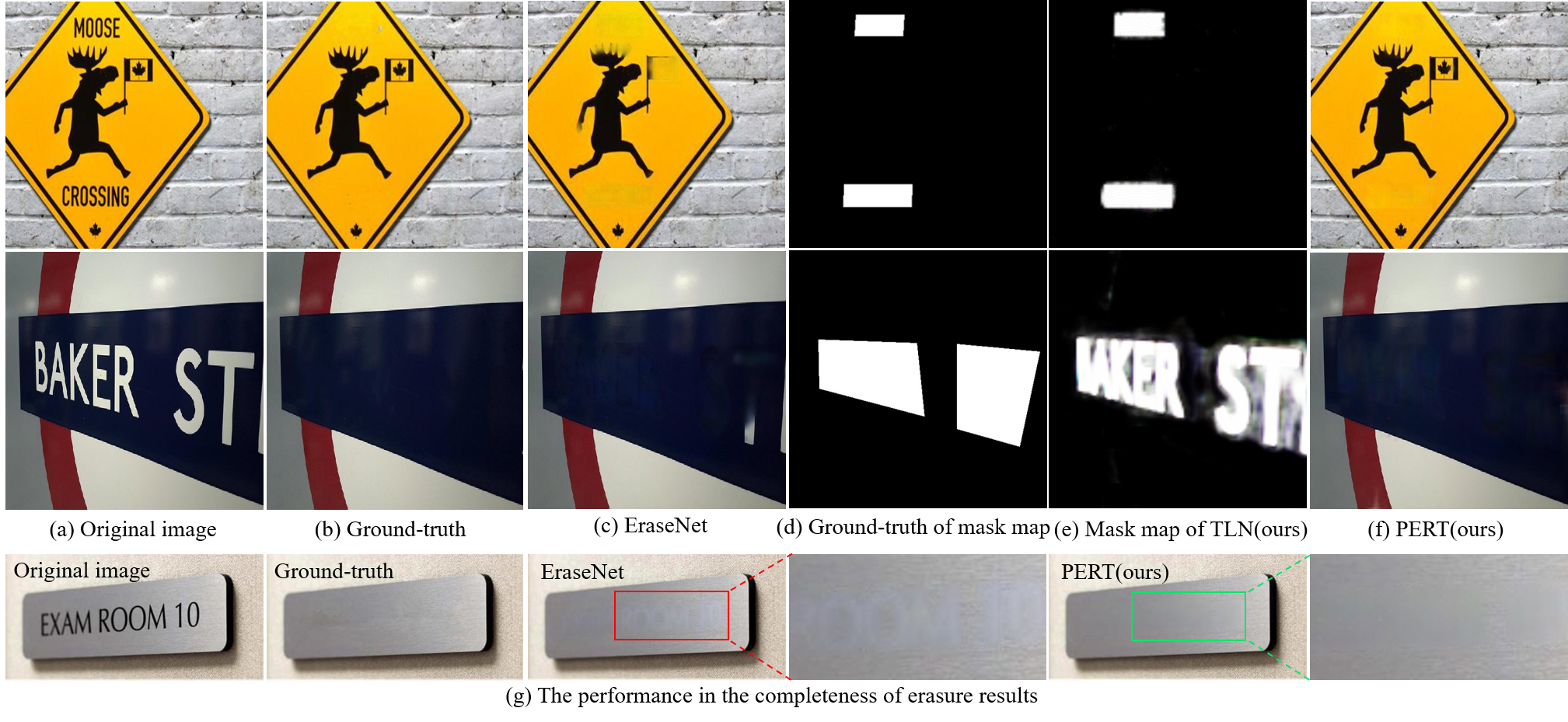}\\
  %\vspace{-1em}
  \caption{The visualization of erasure results on SCUT-EnsText.}
  %\vspace{-2em}
\label{fig:result}
\end{figure*}

We summarize the reasons why RegionMS is more suitable for STR task into two aspects: 1) As the reconstruction learning (Eq. \ref{eq:lg}) on final erasure result $I^{t}_{out}$ aims to learn the stroke-level reconstruction rules, RegionsMS promotes TLN to perceive stroke-level information (Fig. \ref{fig:result} (e)) with only bounding box level annotations (Fig. \ref{fig:result} (d)). On the one hand, for text areas with large character spacing, TLN generates the stroke-level mask map to further reduce the modification of background textures. On the other hand, TLN predicts region-level mask map for the text areas with small character spacing, as the region-level mask map only covers little background textures. Thus, such explicit erasure guidance is able to guide an accurate erasure by implementing targeted and regional modification on stroke-level text regions. 2) RegionMS reduces the learning difficulty in BRN, where texture reconstruction of text-free areas is not considered, resulting in a coarse reconstruction (shown in Fig. \ref{fig:iterative}).

% TLN is able to learn stroke-level information to provide accurate guidance with only bounding box level annotations (detailed in Sec. \ref{sec:regionms}).

\begin{table}[t!p]
\begin{center}
\begin{tabular}{|l|c|c|c|}
\hline
Method & PSNR & MSE & Params \\
\hline
% &  &  &  &  &  & \\
Pix2Pix \cite{isola2017image} &  26.8& 0.0027 &  54.4M\\
MTRNet++ \cite{tursun2020mtrnet++} &  34.55& 0.0004 &  18.7M\\
EraseNet \cite{liu2020erasenet} &  38.32 & \textbf{0.0002}& 19.7M \\
\textbf{PERT} &  \textbf{39.40}& \textbf{0.0002} & \textbf{14.0M}\\
\hline
\end{tabular}
\end{center}
%\vspace{-1em}
\caption{Comparisons between previous methods and proposed PERT on erasure results and parameter size.}
\label{tab:modelsize}
%\vspace{-1em}
\end{table}
\begin{table}[t!p]
\begin{center}
\begin{tabular}{|l|c|c|c|c|}
\hline
 & EraseNet & PERT-1 & PERT-2& PERT-3 \\
\hline
% &  &  &  &  &  & \\
MSSIM & 95.42 & 96.36 & 96.83 & \textbf{96.95}\\
Training (h) & 34.7 & \textbf{11.4} & 14.3 & 16.6 \\
Testing (FPS) & 86 & \textbf{204} & 101 & 71 \\
% Size (Tab.6) & 19.7M & \textbf{14.0M} & \textbf{14.0M} & \textbf{14.0M}\\
\hline
\end{tabular}
\end{center}
%\vspace{-1.1em}
\caption{Comparisons between previous methods and proposed PERT on speed. PERT-i means using i erasing stages.}
\label{tab:speed_table}
%\vspace{-1.5em}
\end{table}

% 2) Instead of modifying all the pixels in the entire image, the explicit erasure guidance provides targeted and regional erasure on only text regions.

\subsubsection{The significance of balanced multi-stage erasure}
\begin{figure}[t!p]
  \centering
  \includegraphics[width=8.5cm]{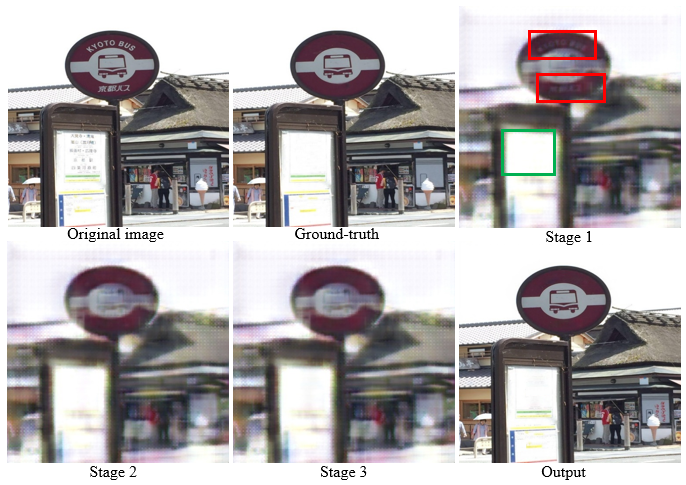}
  %\vspace{-0.5em}
  \caption{The visualization of reconstructed images from BRN in different stages. Output is the erased image from RegionMS in the last stage.}\label{fig:iterative}
  %\vspace{-0.5em}
\end{figure}

By designing a balanced erasure structure and only supervising the last erasing stage, PERT performs a balanced multi-stage erasure. To be specific, for difficult cases (red boxes in Fig. \ref{fig:iterative}), BRN optimizes the erasure result from the previous stage in an equally small-range step. Thus, exhaustive erasure of difficult cases is achieved by progressively erasing. For the relatively easy cases (green box in Fig. \ref{fig:iterative}), PERT tends to achieve a more exhaustive erasure in the early stages.

\subsubsection{The significance in training}

\begin{figure}[t!p]
  \centering
  \includegraphics[width=5.5cm]{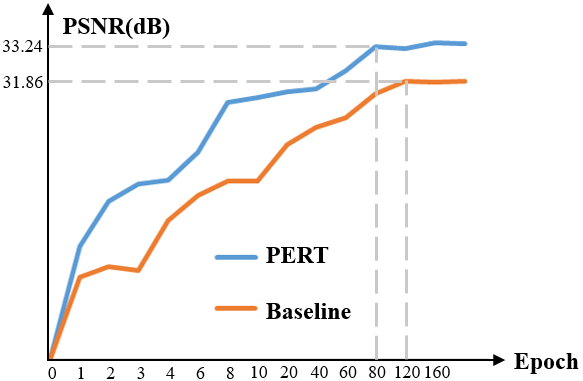}
  \caption{The comparison of convergence in the training stage between PERT and baseline model.}\label{fig:training}
  %\vspace{-1em}
\end{figure}

We visualize the PSNR value on the SCUT-EnsText during training. The baseline model is implemented with adversarial loss and constructed without balanced multi-stage erasure and RegionMS. As shown in Fig. \ref{fig:training}, PERT obtains a better (33.24 vs 31.86) and faster (80 epoch vs 120 epoch) convergence.
% \vspace{-0.5em}
% \subsubsection{The Smoothness On The Edge of Text Region}

% The main concern of RegionMS is the erasure smoothness on the edge of text area. However, such worry is superfluous under our framework. As shown in Fig.\ref{fig:result} (d), the value of pixels in the mask map is smooth on the edge of text area, ensuring the consistency of the erasure result.

% The main concern of RegionMS is the smoothness of transition on the edge of text area. However, such worry is superfluous under our framework.

\subsubsection{Limitation}

% Our method fails in some difficult cases when TLN provides an inaccurate detection result (detecting non-text region or missing detecting text region). However, previous methods \cite{zhang2019ensnet,liu2020erasenet} also fail in these difficult cases. It is worth mention that PERT provides a solution by embedding the latest detection branch \cite{zhang2019look,wang2020contournet,wang2020r}. The detailed visualization is available in the supplementary material.

\begin{figure}[t!p]
 \centering
 \includegraphics[width=5cm]{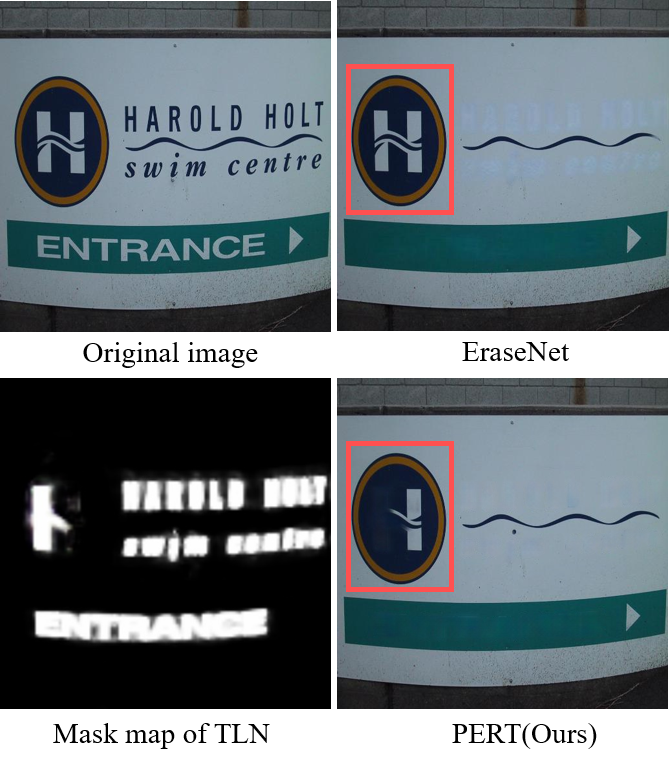}
 %\vspace{-1em}
 \caption{The example of failed cases.}\label{fig:error}
 %\vspace{-1.5em}
\end{figure}

As shown in Fig.\ref{fig:error}, our method fails when TLN provides an inaccurate detection result (detecting non-text region or missing detecting text region). However, this problem also exists in previous methods (EraseNet \cite{liu2020erasenet} for example). It is worth mentioning that our method reduces the impact of this problem to a certain extent (\emph{e.g.} the erasure result of character "H" in Fig.\ref{fig:error}). Furthermore, the proposed PERT provides a solution for this issue by embedding the latest detection branch \cite{zhang2019look,wang2020contournet} to improve the quality of mask map.

\section{Conclusion}
This paper proposes a simple but strong scene text eraser named PERT. Based on the explicit erasure guidance and balanced multi-stage erasure, we qualitatively and quantitatively verify that PERT effectively handles the excessive and inexhaustive erasure problems in STR task. The simplicity of PERT makes it easy to develop new scene text removal models by modifying the existing ones or introducing other network modules. Extensive experiments demonstrate that the proposed method achieves state-of-the-art performance on both synthetic and real-world datasets while maintaining a low complexity. In the future, we will develop this work to the end-to-end text edit task.

{\small
\bibliographystyle{ieee_fullname}
\bibliography{egbib_iccv}
}

\end{document}